\documentclass[10pt,twocolumn,letterpaper]{article}

\usepackage{cvpr}              % To produce the CAMERA-READY version
\usepackage{capt-of}
\usepackage{multirow}

\definecolor{cvprblue}{rgb}{0.21,0.49,0.74}
\usepackage[pagebackref,breaklinks,colorlinks,allcolors=cvprblue]{hyperref}

\def\confName{CVPR}
\def\confYear{2027}

\title{Ego4Gen: Generating Geometry-Grounded Motion Tokens as a World-Action Interface}

\author{%
  {\fontsize{10.95}{13.6}\selectfont
   Yang Liu\textsuperscript{1}, Chao Ye\textsuperscript{1,*},
   Jiahao Zhang\textsuperscript{1}, Zeyuan Hou\textsuperscript{1},
   Ruijie Fang\textsuperscript{1}, Xiyin Ni\textsuperscript{1},
   Xinyang Ren\textsuperscript{2}, Weiyang Lin\textsuperscript{1}}\\[3pt]
  {\normalsize \textsuperscript{1}\,Harbin Institute of Technology\qquad
               \textsuperscript{2}\,Yunnan Minzu University}\\
  {\small \textsuperscript{*}\,Corresponding author.}%
}
\hypersetup{pdfauthor={Yang Liu, Chao Ye, Jiahao Zhang, Zeyuan Hou, Ruijie Fang, Xiyin Ni, Xinyang Ren, Weiyang Lin}}

\begin{document}

% \maketitle

\makeatletter
\twocolumn[{
    \@maketitle
    \begin{center}
        \includegraphics[width=1.0\textwidth]{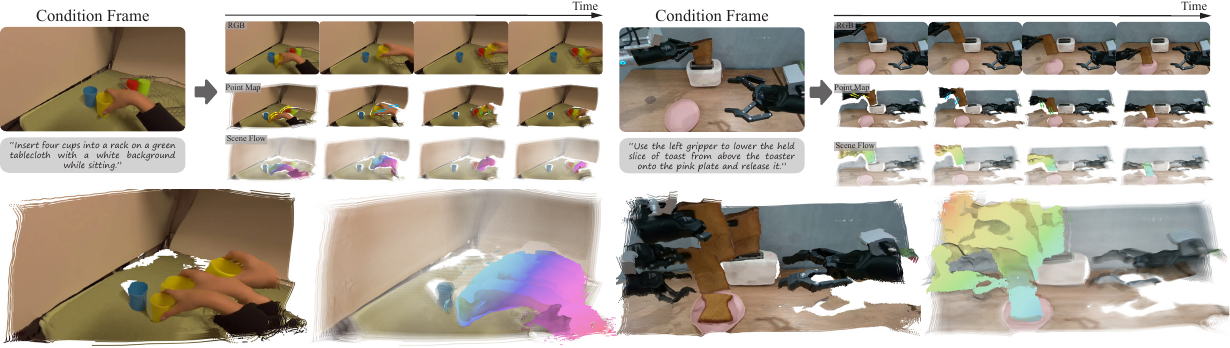}
        \captionof{figure}{Overview of Ego4Gen. Given an initial egocentric RGB image and a text prompt, Ego4Gen generates consistent future RGB frames, point maps, and dense scene flow, which together represent the 4D future.}
        \label{overview}
    \end{center}
    \vspace{1pt}
}]
\makeatother

\begin{abstract}
We present Ego4Gen (Egocentric 4D Future Generation), a world model for 4D future generation in robotic manipulation. Given a single egocentric image and a text prompt, Ego4Gen jointly generates temporally aligned RGB frames, dense point maps, and dense scene flow, explicitly predicting the evolution of scene appearance, spatial structure, and motion. Built on Cosmos3-Nano, Ego4Gen extends the pretrained generative backbone through modality-routed LoRA and asymmetric attention, using geometric pseudo-labels from MotionCrafter to learn to predict future geometry and motion latents. These latents can either be decoded into explicit 3D structure and motion or mapped to robot actions. To this end, we also introduce FlowInverse, a lightweight inverse dynamics network trained in two stages to translate predicted 4D latents into robot actions. Experiments show that these geometry and motion latents outperform the evaluated latent-action and video-latent baselines in out-of-distribution settings. Closed-loop policy evaluations further demonstrate their utility for translating predicted 4D futures into robot control. Together, these results support geometry-grounded motion tokens as a world-action interface, connecting 4D future generation with robotic manipulation and suggesting a path toward reduced reliance on action-labeled data. The project page is available at \url{https://boominheart.github.io/Ego4Gen}.
\end{abstract}
\section{Introduction}
\label{sec:intro}

Visual world models offer a new way to learn interaction dynamics from large-scale video data for robotic manipulation. By predicting how a task may unfold, they support action planning and generation \cite{NEURIPS2023_1d5b9233,ye2026rss}. However, RGB videos encode task-relevant state changes together with appearance factors such as color, texture, lighting, and background. Even scenes with similar spatial relationships and motions can differ substantially in pixel appearance, creating appearance shifts that a controller must accommodate. The representation connecting future prediction to action execution should therefore preserve control-relevant information while reducing interference from task-irrelevant visual variation.

Recent works explore latent-space interfaces between visual priors and robot actions. Some approaches perform planning with predicted video latent representations \cite{assran2025vjepa2}, while others use latent action models (LAMs) that compress inter-frame changes into compact action representations \cite{pmlr-v306-gao26q,BuQ-RSS-25}. However, these compact representations may still retain visual information unrelated to robot control. Prior analysis shows that latent action representations can encode exogenous noise alongside action-induced changes \cite{NEURIPS2025_d7b3cef7}. This suggests that compactness alone is insufficient to ensure robustness to appearance shifts.

Geometry offers a promising direction for reducing the influence of task-irrelevant visual appearance. Recent studies have explored mapping predicted geometry and motion to robot actions. Some methods use explicitly constructed geometric representations \cite{pmlr-v270-xu25a,Zhen_2025_ICCV,zhi20253dflowactionlearningcrossembodimentmanipulation}, while others extract features from within world models \cite{rynnworld4d,li_hydra-0_2026}. The former introduces additional computation to construct and process explicit representations, while the latter can still be affected by shifts in the visual input distribution. We therefore seek an interface that contains as little task-irrelevant information as possible, motivating our focus on geometry-motion representations.

Our experiments show that the geometry-motion latents of MotionCrafter's 4D VAE \cite{Zhu_2026_CVPR} are well suited to this role: they support both explicit 4D scene reconstruction and robust mappings to robot actions. Building on this finding, we develop Ego4Gen upon Cosmos3-Nano \cite{nvidia2026cosmos3}. Given a single egocentric image and a text prompt, Ego4Gen jointly predicts future RGB video and temporally aligned geometry-motion latents, which can be decoded into dense point maps and dense scene flow. To connect these predictions to robot control, we introduce FlowInverse, a lightweight inverse dynamics network that enables closed-loop robot control with Ego4Gen through two-stage training: learning from real videos and adapting to predicted representations. We further show that adapting the world model with action-unlabeled videos can improve generalization in closed-loop control while keeping FlowInverse fixed, highlighting the potential to benefit from additional video data without additional action supervision.

Overall, our contributions are as follows:
\begin{itemize}
    \item We validate geometry-motion latents as a world-action interface. Through comparisons with latent action and video representations, we show that MotionCrafter's 4D VAE latents support both explicit 3D structure reconstruction and robot action prediction, with strong robustness under the out-of-distribution conditions.

    \item We introduce Ego4Gen, a 4D world model that jointly predicts temporally aligned future RGB frames, dense point maps, and dense scene flow from a single egocentric image and a text prompt.

    \item We introduce FlowInverse, a lightweight inverse dynamics network that maps Ego4Gen's geometry-motion latents to robot actions for closed-loop control. We also demonstrate that world-model adaptation with action-unlabeled videos can improve downstream policy performance, suggesting a way to reduce reliance on action-labeled data.
\end{itemize}

\begin{figure*}[t]\centering
	\includegraphics[width=1.0\textwidth, clip=true]{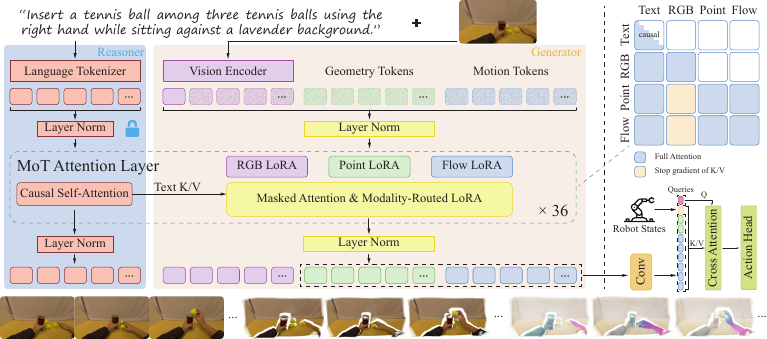}  % left bottom right top
	\caption{Framework of Ego4Gen. Built on the image-to-video (I2V) pathway of Cosmos3-Nano, Ego4Gen augments the generator with geometry and motion modalities, adapts the shared backbone using modality-routed LoRA, and jointly denoises the three modalities through asymmetric attention. The resulting geometry and motion latents can further support closed-loop action-policy generation.}\label{framework}
\vspace{-0.4cm}
\end{figure*}
\section{Related Works}

\subsection{4D World Action Models}

Visual world action models have made substantial progress in transferring knowledge from large-scale video data to robotic manipulation by combining visual generation and action modeling \cite{nvidia2026cosmos3,sun2026riemann10embodiedworldaction}. Building on these advances, recent work incorporates geometry and motion to better support robot control. Some methods jointly model depth and other geometric modalities to provide explicit spatial information \cite{guo2026unified4dworldaction,Zhen_2025_ICCV}, while others introduce motion cues through optical flow prediction \cite{rynnworld4d} or action flow conditioning \cite{li_hydra-0_2026}.

We seek to introduce more explicit temporal and motion representations into world-model outputs. However, existing work represents 4D scene dynamics in different ways: some methods use dynamic point maps to describe evolving 3D geometry \cite{mi2026one4d,liu2026pixelsvideopriors4d}, while others represent motion in the image plane through optical flow or action flow \cite{rynnworld4d,li_hydra-0_2026}. We adopt dense 3D scene flow \cite{lu2026track4world,Zhu_2026_CVPR}, which directly describes the displacement of scene points in 3D space and provides an intuitive physical description of motion.

\subsection{World-Action Interface}

The world-action interface which connects world model prediction to action execution is central to world action models. Existing approaches employ several forms of coupling. Some jointly model video and actions within a unified architecture \cite{nvidia2026cosmos3,sun2026riemann10embodiedworldaction}, while others extract internal world-model features for action prediction \cite{rynnworld4d,li_hydra-0_2026}. Hierarchical approaches first predict the future and then translate the predictions into actions using a separate controller \cite{NEURIPS2023_1d5b9233,Zhen_2025_ICCV}.

Research on world-action interfaces also explores representations suited to action prediction. LAMs learn action representations by compressing changes between frames, providing a bridge to learning from videos without action labels \cite{pmlr-v306-gao26q,BuQ-RSS-25}. However, these representations can also encode visual changes unrelated to control \cite{NEURIPS2025_d7b3cef7}. Other methods connect future generation to robot control through explicit object motion predictions \cite{pmlr-v270-xu25a}, involving the unnecessary generation, processing, and subsequent encoding of motion representations.

We use geometry-motion latents from a pretrained 4D VAE \cite{Zhu_2026_CVPR} as the world-action interface. We use a generative model to predict future latents and a lightweight inverse dynamics network to convert them into robot actions. This interface preserves the ability to reconstruct explicit 3D geometry and motion while directly supporting robot control in latent space.
\section{Method}

% \begin{figure*}[t]\centering
% 	\includegraphics[width=1.0\textwidth, clip=true]{pics/fig2.pdf}  % left bottom right top
% 	\caption{Framework of Ego4Gen. Built on the image-to-video (I2V) pathway of Cosmos3-Nano, Ego4Gen augments the generator with geometry and motion modalities, adapts the shared backbone using modality-routed LoRA, and jointly denoises the three modalities through asymmetric attention. The resulting geometry and motion latents can further support closed-loop action-policy generation.}\label{framework}
% \end{figure*}

\subsection{Problem Formulation and Preliminaries}\label{pfp}

Ego4Gen aims to jointly generate a 4D future from a single initial egocentric image and a text prompt. Given an initial observation $I_0$ and a text prompt $c$, Ego4Gen jointly predicts RGB frames $R$, dense point maps $P$, and dense scene flow $F$ over the next four seconds. We refer to these three streams as the vision ($V$), geometry ($G$), and motion ($M$) modalities, respectively, and model their conditional distribution as
\begin{align}
p_{\theta}(R,P,F\mid I_0,c),
\end{align}
where $R=\{I_i\}_{i=0}^{T_V-1}\in\mathbb{R}^{T_V\times 3\times H\times W}$ denotes the future RGB frames including the initial frame $I_0$, while $P=\{P_j\}_{j=0}^{T_{GM}-1}$ and $F=\{F_j\}_{j=0}^{T_{GM}-1}$, both in $\mathbb{R}^{T_{GM}\times 3\times H\times W}$, denote the future point maps and scene flow, respectively. Each point map assigns a 3D location to each pixel, while scene flow represents the associated 3D forward displacement, following the coordinate and scale conventions of MotionCrafter~\cite{Zhu_2026_CVPR}. The vision and geometry-motion sequences cover the same prediction window but are sampled at different rates: the vision sequence contains $T_V=96$ frames at 24 FPS, whereas geometry-motion sequences each contain $T_{GM}=48$ frames at 12 FPS.

We choose Cosmos3-Nano \cite{nvidia2026cosmos3} as the pretrained backbone and extend its Mixture-of-Transformers (MoT) architecture to jointly generate RGB, point map, and scene flow latents conditioned on the input image and text prompt. For vision, we retain the latent representation produced by the Wan2.2 VAE \cite{wan2025}. For geometry and motion, we adopt the latent representations provided by MotionCrafter.

\subsection{Unified Multimodal Generation}\label{UMG}

As shown in \cref{framework}, we adapt the MoT architecture of Cosmos3-Nano to the task defined in \cref{pfp}. We retain its text reasoner and image-to-video (I2V) generation pathway, while disabling audio and action generation. We then freeze the pretrained backbone weights and introduce geometry and motion branches to generate dense point maps and scene flow, respectively. The three continuous modalities share the pretrained backbone, use modality-routed LoRA for modality-specific adaptation, and are jointly denoised with asymmetric attention.

\subsubsection{Spatiotemporal Alignment}

During training, the text prompt is tokenized and processed by the frozen reasoner. RGB frames are encoded by the Wan2.2 VAE, while geometry and motion latent targets are obtained from the downsampled frames using a frozen MotionCrafter predictor. We duplicate the initial frame $I_0$ as the first frame of the prediction target and use its timestamp as the start of the four-second prediction horizon. This allows the model to predict geometry and motion for the conditioning frame as well, yielding a complete 4D representation from the initial timestamp onward.

For the vision modality, the conditioning frame $I_0$ occupies a separate temporal slot in the latent space, while the future frames supervised during training are temporally compressed by a factor of four following Wan2.2, resulting in 25 vision slots in total. MotionCrafter's geometry and motion latents have no temporal compression and use a different spatial compression factor. We therefore apply modality-specific spatiotemporal patchification to the geometry and motion latents to align the three latent token spaces. This results in 25 vision slots and 24 slots for each of the geometry and motion modalities, with spatial tokens of the same shape in every slot.

\begin{figure}[htbp]\centering
	\includegraphics[width=1.0\linewidth, clip=true]{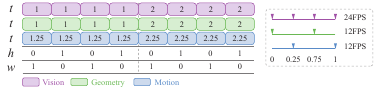}  % left bottom right top
	\caption{Spatiotemporal position coordinates. The three modalities share a spatial token grid. Motion, defined between consecutive geometry frames, has a temporal offset of $0.25$ from the initial time.}\label{time}
\end{figure}

As illustrated in \cref{time}, we use the 3D MRoPE adopted by Cosmos3, together with its absolute temporal modulation. In our setting, the base temporal rate is $\mathrm{TPS}_{\mathrm{base}}=24/4=6$, so a unit increment on the temporal axis corresponds to $1/6$ second. Within each temporal slot, the height and width coordinates are assigned according to the spatial token grid. Since all three modalities use the same spatial resolution, geometry and motion tokens share the same spatial position indices as the vision tokens. The arrows in the right part of \cref{time} indicate the temporal correspondence between modality frames assigned to the same slot. Because each motion frame represents the 3D displacement between consecutive geometry frames, we assign the motion modality an absolute temporal offset of $0.25$ from the initial time. Since the initial frame $I_0$ is duplicated as the first frame of the target sequence, the temporal coordinates are
\begin{equation}
\begin{aligned}
\boldsymbol{\tau}^{V} &= (0,0,1,\ldots,23),\\[-1pt]
\boldsymbol{\tau}^{G} &= (0,1,\ldots,23),\\[-1pt]
\boldsymbol{\tau}^{M} &= (0.25,1.25,\ldots,23.25)
\end{aligned}
\end{equation}
for the vision, geometry, and motion modalities, respectively.

After denoising, the generated latent sequences are decoded into RGB frames, point maps, and scene flow using their corresponding VAE decoders:
\begin{equation}
\hat R=D_R(\hat z^V),\;
\hat P=D_P(\hat z^G),\;
\hat F=D_F(\hat z^M,\hat z^G).
\end{equation}
Following MotionCrafter, we set the final scene-flow frame to zero, since no subsequent geometry frame exists to define its forward displacement.

\subsubsection{Modality-Routed LoRA \& Asymmetric Attention}

To adapt Cosmos3-Nano to multimodal generation while preserving its pretrained capabilities, we use modality-routed LoRA to adapt its generator. Specifically, we maintain three modality-specific sets of LoRA parameters for the vision, geometry, and motion modalities, and route each token according to its modality label. The routed LoRA updates are inserted into the attention projections $s\in\{Q,K,V,O\}$, corresponding to independent low-rank updates for each modality $m\in\{V,G,M\}$:
\begin{equation}
\widetilde W_s^{(m)} = W_s+\frac{\alpha_m}{r_m}B_s^{(m)}A_s^{(m)}.
\end{equation}
Text tokens use the frozen base parameters, while the input and output heads for geometry and motion are optimized together with the LoRA parameters. This design allows us to use modality-specific LoRA configurations, thereby reducing disruption to the pretrained priors when adapting the model with relatively limited data.

Modality-specific parameterization alone does not determine how information flows between modalities. We therefore extend the attention mask of Cosmos3 to introduce asymmetric attention. As illustrated in the upper-right of \cref{framework}, the original causal attention pattern is preserved for text tokens in the reasoner branch. In the generator branch, RGB tokens attend only to text and RGB tokens, whereas point and flow tokens attend to all modalities. This design allows geometry and motion to exploit RGB context during denoising while shielding RGB generation from geometric interference.

To further reduce the interference of geometry supervision with vision generation, point and flow tokens use RGB keys and values with detached gradients when computing attention. Accordingly, attention is computed for the three modalities as follows:
\begin{equation}
\label{eq:asymmetric_attention}
\begin{aligned}
O_V
&= \operatorname{Attn}\!\left(
Q_V,\,[K_T;K_V],\,[V_T;V_V]\right),\\[-1pt]
O_{G/M}
&= \operatorname{Attn}\!\Big(
Q_{G/M},\,
[K_T;\operatorname{sg}(K_V);K_G;K_M],\\[-1pt]
&\hspace{4.4em}
[V_T;\operatorname{sg}(V_V);V_G;V_M]\Big).
\end{aligned}
\end{equation}
Where $\operatorname{sg}(\cdot)$ denotes the stop-gradient operation. This establishes a one-directional conditioning pathway from RGB to geometry and motion, while enabling the three continuous modalities to be denoised jointly.

We train Ego4Gen with a multimodal rectified-flow matching objective. For each modality $m\in\{V,G,M\}$, the model predicts a velocity field $\hat{v}_m$ toward the corresponding clean latent $z_m^0$. The target velocity is defined as $v_m^{*}=\epsilon_m-z_m^0$, where $\epsilon_m$ is Gaussian noise. We compute an MSE loss for each modality and combine the three losses as

\begin{equation}
\begin{aligned}
\mathcal{L} &= \mathcal{L}_V + \mathcal{L}_G + \mathcal{L}_M, \\
\mathcal{L}_m &= \operatorname{MSE} \left(\hat{v}_m,\,\epsilon_m-z_m^0\right).
\end{aligned}
\end{equation}

\subsection{FlowInverse: Latent-to-Action Translation}

To convert the predicted geometry-motion latent representation into robot control commands, we introduce FlowInverse, a lightweight inverse dynamics network. Given the current robot states and the future geometry-motion latents, FlowInverse directly predicts the corresponding robot action increments.

As illustrated in the lower-right of \cref{framework}, we process the geometry and motion latents with two separate convolutional encoders and add spatiotemporal positional encodings and modality embeddings to the resulting tokens. We then use action queries to perform cross-attention over tokens from the two latent streams and the proprioceptive state. This design allows the action queries to incorporate information from the robot's proprioceptive state and selectively read action-relevant features from the spatially structured geometry representation.

For output supervision, we use the joint-action increments and gripper states along the future trajectory as the primary supervision signals, while using end-effector displacement prediction as an auxiliary supervision signal. The network is trained with a weighted MSE regression loss:
\begin{equation}
\mathcal{L} = \lambda_{\mathrm{joint}}\mathcal{L}_{\mathrm{joint}} + \lambda_{\mathrm{gripper}}\mathcal{L}_{\mathrm{gripper}} + \lambda_{\mathrm{ee}}\mathcal{L}_{\mathrm{ee}}.
\end{equation}

\subsection{Datasets \& Training}

During Ego4Gen pretraining, we use EgoDex \cite{hoque2026egodex}, AgiBot \cite{agibotworld2026}, and clean-task trajectories from the ALOHA-AgileX robot in RoboTwin 2.0 \cite{chen2025robotwin}. All data are organized into four-second training windows. We use modality-routed LoRA adapters with $(r,\alpha)=(32,64)$ for the vision branch and $(r,\alpha)=(64,128)$ for both the geometry and motion branches. We train the model on 8 NVIDIA RTX PRO 6000 GPUs with a global batch size of 24 for 40K optimizer steps, using a warmup-stable-decay (WSD) learning-rate schedule.

FlowInverse is trained in two stages. In the first stage, we pretrain it on the 50 clean tasks in RoboTwin 2.0, where geometry-motion latents are extracted from expert videos using a frozen MotionCrafter model. In the second stage, we fine-tune a separate model for each target task using paired data consisting of Ego4Gen generated geometry-motion latents and the corresponding action trajectories. Ego4Gen remains frozen throughout FlowInverse training. Both stages use clean-scene data, while randomized scenes are reserved for evaluation.
\section{Experiments}

\subsection{A Versatile Latent Representation}\label{versatile_representation}

As described in \cref{UMG}, we adopt the geometry-motion latent representation defined by MotionCrafter's 4D VAE, which can be decoded directly into point maps and scene flow. Here, we investigate its use for direct robot control.

We compare the joint geometry-motion (GM) latent representation with a motion-only (M-only) variant. Our baselines include latent actions from the LAMs of DreamDojo \cite{pmlr-v306-gao26q} and UniVLA \cite{BuQ-RSS-25}, as well as video latents from the Wan 2.2 VAE. All representations are extracted from future video segments of expert demonstrations using their respective frozen encoders. We use representation-specific adapters with comparable parameter counts to connect these latents to the FlowInverse architecture, with independently trained weights for each representation.

We evaluate on RoboTwin 2.0 \cite{chen2025robotwin}, selecting Place Empty Cup and Pick Diverse Bottles as the target tasks. For each representation, we first pretrain FlowInverse on clean demonstrations from the remaining 48 tasks, excluding both target tasks. Then we fine-tune a separate policy for each target task using its clean demonstrations. We measure both success rate and offline action prediction error under the Easy and Hard settings. During execution evaluation, the conditioning latents are extracted from videos of an expert policy performing the task in the corresponding evaluation scene. The results are reported in \cref{latent_comparison}.

\begin{table}[t]
    \centering
    \caption{Task success rates and action prediction errors with different representations. Each cell reports success rate (\%, $\uparrow$) / MSE ($10^{-3}\,\mathrm{rad}^{2}$, $\downarrow$). GM and M-only denote MotionCrafter's joint geometry-motion and motion-only latents, respectively.}
    \label{latent_comparison}
    \small
    \setlength{\tabcolsep}{2.5pt}
    \renewcommand{\arraystretch}{1.0}
    \begin{tabular}{@{}lcccc@{}}
        \toprule
        \multirow{2}{*}{Representation}
            & \multicolumn{2}{c}{Place Empty Cup}
            & \multicolumn{2}{c}{Pick Diverse Bottles} \\
        \cmidrule(lr){2-3}\cmidrule(lr){4-5}
            & Easy & Hard & Easy & Hard \\
        \midrule
        GM
            & \underline{82}\hspace{2pt}/\hspace{2pt}\underline{0.66}
            & \underline{51}\hspace{2pt}/\hspace{2pt}\underline{1.72}
            & \textbf{46}\hspace{2pt}/\hspace{2pt}5.18
            & \textbf{37}\hspace{2pt}/\hspace{2pt}\textbf{4.04} \\
        M-only
            & 60\hspace{2pt}/\hspace{2pt}0.73
            & \textbf{55}\hspace{2pt}/\hspace{2pt}\textbf{1.60}
            & 31\hspace{2pt}/\hspace{2pt}5.74
            & \underline{26}\hspace{2pt}/\hspace{2pt}\underline{4.67} \\
        \midrule
        DreamDojo~\cite{pmlr-v306-gao26q}
            & \textbf{88}\hspace{2pt}/\hspace{2pt}\textbf{0.56}
            & 0\hspace{2pt}/\hspace{2pt}12.49
            & \underline{35}\hspace{2pt}/\hspace{2pt}\textbf{4.24}
            & 0\hspace{2pt}/\hspace{2pt}38.91 \\
        UniVLA~\cite{BuQ-RSS-25}
            & 51\hspace{2pt}/\hspace{2pt}0.99
            & 0\hspace{2pt}/\hspace{2pt}15.96
            & 12\hspace{2pt}/\hspace{2pt}7.08
            & 4\hspace{2pt}/\hspace{2pt}12.74 \\
        Wan 2.2~\cite{wan2025}
            & 16\hspace{2pt}/\hspace{2pt}1.96
            & 0\hspace{2pt}/\hspace{2pt}38.01
            & 31\hspace{2pt}/\hspace{2pt}\underline{4.37}
            & 0\hspace{2pt}/\hspace{2pt}100.46 \\
        \bottomrule
    \end{tabular}
\vspace{-0.4cm}
\end{table}

The joint geometry-motion representation achieves success rates of 51\% and 37\% on Place Empty Cup and Pick Diverse Bottles under the Hard setting, respectively, substantially outperforming the LAM and video-latent baselines. The lower offline action prediction errors further support the advantage of geometry-motion representations under distribution shift. Meanwhile, DreamDojo achieves the highest success rate on Place Empty Cup and the lowest action prediction errors on both tasks under the Easy setting, while the motion-only variant performs best on Place Empty Cup under the Hard setting. Nevertheless, the joint representation achieves stronger overall performance. These results support the value of geometry-motion latents as a robust interface for action decoding, beyond their use for explicit 3D decoding.

\subsection{4D Generation}

\begin{figure*}[t]\centering
	\includegraphics[width=1.0\textwidth, clip=true]{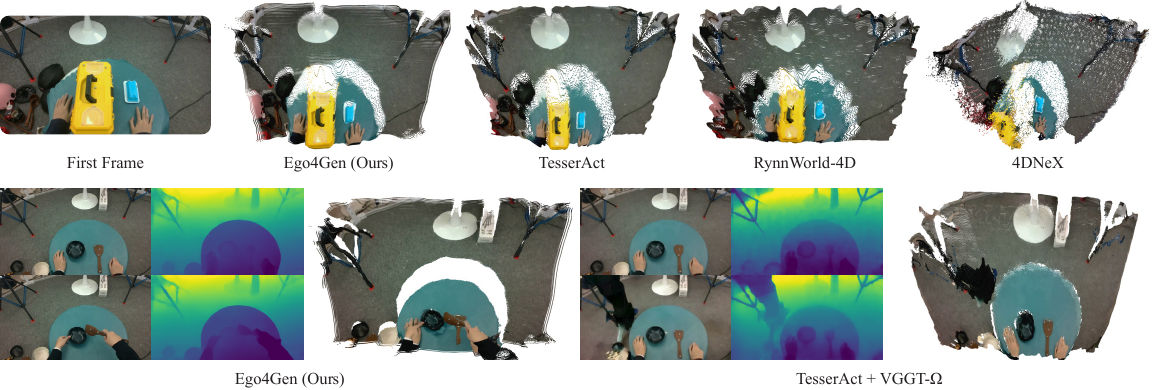}  % left bottom right top
	\caption{Visualization of generated point maps. Top: first-frame reconstructions from different models. Bottom: point maps fused across multiple frames, with the same edge filtering applied to all methods. Ego4Gen produces sharper, better-defined point map boundaries.}\label{4d_generation}
\vspace{-0.4cm}
\end{figure*}

\begin{table}[t]
    \centering
    \caption{4D generation on TACO over a two-second horizon.}
    \label{wm2s}
    \setlength{\tabcolsep}{0.7pt}
    \begin{tabular}{@{}lcccccc@{}}
        \toprule
        \multirow{2}{*}{Method}
            & \multicolumn{2}{c}{RGB}
            & \multicolumn{2}{c}{Depth}
            & \multicolumn{2}{c}{Points (\%)} \\
        \cmidrule(lr){2-3}
        \cmidrule(lr){4-5}
        \cmidrule(lr){6-7}
            & PSNR$\uparrow$
            & LPIPS$\downarrow$
            & AbsRel$\downarrow$
            & $\delta_1{\uparrow}$
            & $\mathrm{Rel}_p{\downarrow}$
            & $\delta_p{\uparrow}$ \\
        \midrule
        Wan2.2
            & 14.93 & 0.440
            & -- & -- & -- & -- \\
        Cosmos3-Nano
            & 15.89 & \underline{0.398}
            & -- & -- & -- & -- \\
        RynnWorld-4D
            & \textbf{16.27} & 0.413
            & 0.174 & 0.726 & -- & -- \\
        4DNeX
            & 15.04 & 0.465
            & -- & --
            & \underline{30.34} & \underline{50.80} \\
        TesserAct
            & 15.71 & 0.423
            & \underline{0.144} & \underline{0.798}
            & -- & -- \\
        \midrule
        Ego4Gen
            & \underline{16.06} & \textbf{0.394}
            & \textbf{0.140} & \textbf{0.807}
            & \textbf{23.21} & \textbf{73.62} \\
        \bottomrule
    \end{tabular}
\vspace{-0.4cm}
\end{table}

We evaluate Ego4Gen's 4D generation quality on TACO \cite{Liu_2024_CVPR}, which is not included in our training data. It is an egocentric bimanual manipulation dataset that provides RGB videos, depth observations, camera intrinsics, and per-frame extrinsics. We use the provided depth and camera calibration to construct reference depth maps and point maps, and compare Ego4Gen with Cosmos3-Nano \cite{nvidia2026cosmos3}, Wan2.2-TI2V-5B \cite{wan2025}, RynnWorld-4D \cite{rynnworld4d}, 4DNeX \cite{chen20254dnex}, and TesserAct \cite{Zhen_2025_ICCV}. All models are conditioned on the same initial RGB image and text prompt. Following its official setup, RynnWorld-4D additionally uses an initial depth map estimated from the input image by Depth Anything 3 \cite{depthanything3}. To account for differences in native frame rates and prediction horizons, we evaluate at timestamps shared by the models in each comparison. Since 4DNeX generates only two-second sequences, \cref{wm2s} reports results over a two-second horizon for all models, while \cref{wm4s} reports four-second results for models supporting this longer horizon, which can further assess the models' long-horizon generation capabilities.

For RGB generation, we report PSNR and LPIPS against the ground-truth future videos. To accommodate differences in output modalities, we evaluate geometric quality in terms of both depth and point maps, reporting AbsRel and threshold accuracy $\delta_1$ at a threshold of $1.25$ for depth, and $\mathrm{Rel}_p$ and $\delta_p$ for point maps. Here, $\delta_p$ denotes the fraction of valid points with a relative 3D error below $0.25$. Since Ego4Gen does not include a camera-parameter prediction branch, we use TACO's camera intrinsics and per-frame extrinsics to convert its predicted point maps into depth maps after coordinate and scale alignment, enabling comparison with models that predict depth directly.

\begin{table}[t]
    \centering
    \caption{4D generation on TACO over a four-second horizon.}
    \label{wm4s}
    \setlength{\tabcolsep}{0.7pt}
    \begin{tabular}{@{}lcccccc@{}}
        \toprule
        \multirow{2}{*}{Method}
            & \multicolumn{2}{c}{RGB}
            & \multicolumn{2}{c}{Depth}
            & \multicolumn{2}{c}{Points (\%)} \\
        \cmidrule(lr){2-3}
        \cmidrule(lr){4-5}
        \cmidrule(lr){6-7}
            & PSNR$\uparrow$
            & LPIPS$\downarrow$
            & AbsRel$\downarrow$
            & $\delta_1{\uparrow}$
            & $\mathrm{Rel}_p{\downarrow}$
            & $\delta_p{\uparrow}$ \\
        \midrule
        Wan2.2
            & 13.69 & 0.513
            & -- & --
            & -- & -- \\
        Cosmos3-Nano
            & 14.69 & \underline{0.461}
            & -- & --
            & -- & -- \\
        RynnWorld-4D
            & \underline{14.79} & 0.493
            & 0.184 & 0.697
            & -- & -- \\
        TesserAct
            & 14.35 & 0.491
            & \underline{0.160} & \underline{0.763}
            & -- & -- \\
        \midrule
        Ego4Gen
            & \textbf{14.83} & \textbf{0.456}
            & \textbf{0.153} & \textbf{0.773}
            & \textbf{26.77} & \textbf{67.47} \\
        \bottomrule
    \end{tabular}
\end{table}

\begin{figure}[t]\centering
	\includegraphics[width=1.0\linewidth, clip=true]{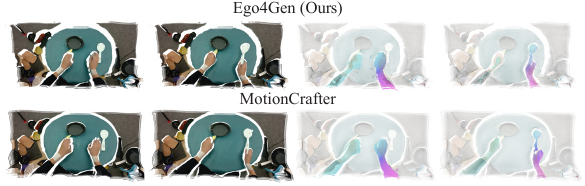}  % left bottom right top
	\caption{Visualization of generated dense scene flow. For comparison, MotionCrafter is run on the same RGB video generated by Ego4Gen.}\label{motion_visualization}
\vspace{-0.4cm}
\end{figure}

For the motion modality, ground-truth 3D motion is unavailable. We therefore use VGGT-$\Omega$ \cite{wang26vggt-omega} and AllTracker \cite{harley2025alltracker} to construct 3D motion pseudo-references from the RGB videos generated by Ego4Gen. We estimate a scale factor for each video using the first-frame depth and camera intrinsics provided by TACO, and express both the pseudo-references and the aligned predictions in meters. We compare Ego4Gen's jointly generated geometry and motion with MotionCrafter's reconstructions from the same RGB videos. Since most pixels in these videos correspond to static regions, we restrict motion evaluation to valid dynamic points to prevent static regions from dominating the aggregate metrics. We evaluate 3D motion consistency at these points between consecutive frames sampled at 12 Hz. Motion magnitudes in our egocentric setting are relatively small, with reference displacements typically on the order of centimeters. We therefore report endpoint error (EPE, in centimeters) and $\mathrm{APD}_{0.01}$, where the latter measures the percentage of valid dynamic points whose predicted 3D displacements have an endpoint error of at most 1~cm relative to the scale-calibrated pseudo-references. Results are reported in \cref{wmflow}.

\begin{table}[t]
    \centering
    \caption{Geometry and motion consistency with VGGT-$\Omega$ + AllTracker pseudo-references on the same Ego4Gen-generated RGB videos.}
    \label{wmflow}
    \setlength{\tabcolsep}{0.7pt}
    \begin{tabular}{@{}lcccc@{}}
        \toprule
        \multirow{2}{*}{Method}
            & \multicolumn{2}{c}{Points (\%)}
            & \multicolumn{2}{c}{Motion} \\
        \cmidrule(lr){2-3}
        \cmidrule(lr){4-5}
            & $\mathrm{Rel}_p{\downarrow}$
            & $\delta_p{\uparrow}$
            & EPE (cm)$\downarrow$
            & $\mathrm{APD}_{0.01}$ (\%)$\uparrow$ \\
        \midrule
        MotionCrafter
            & 11.04 & 92.41
            & \textbf{1.430} & \textbf{48.67} \\
        \midrule
        Ego4Gen
            & \textbf{10.77} & \textbf{92.69}
            & 1.452 & 47.22 \\
        \bottomrule
    \end{tabular}
\vspace{-0.4cm}
\end{table}

We also present qualitative comparisons of first-frame point map reconstructions and multi-frame fusion results for dynamic scenes in \cref{4d_generation}. For TesserAct, we reconstruct point clouds using camera parameters estimated by VGGT-$\Omega$. \cref{motion_visualization} further compares the point maps and scene flow generated by Ego4Gen with MotionCrafter's reconstructions, obtained from the same RGB video generated by Ego4Gen.

These results demonstrate Ego4Gen's leading performance in 4D generation. Its output quality closely matches that of MotionCrafter's reconstructions, supporting its use in closed-loop robot control.

\subsection{Closed-Loop Robot Control}\label{CLRC}

\begin{table*}[t]
    \centering
    \caption{Closed-loop success rates (\%, $\uparrow$) on five RoboTwin 2.0 tasks. Each cell reports Easy / Hard. Mixed uses clean and randomized videos without action labels for world-model post-training.}
    \label{closed_loop}
    \setlength{\tabcolsep}{4pt}
    \begin{tabular}{@{}lcccccc@{}}
        \toprule
        Method
            & \shortstack{Grab\\Roller}
            & \shortstack{Place Empty\\Cup}
            & \shortstack{Pick Diverse\\Bottles}
            & \shortstack{Place\\Shoe}
            & \shortstack{Place Bread\\Basket}
            & Mean \\
        \midrule
        DP
            & \textbf{100} / 3
            & 15 / 0
            & 8 / 1
            & 20 / 0
            & 9 / 0
            & 30.4 / 0.8 \\
        TurboVLA
            & 93 / 3
            & \underline{88} / 1
            & 40 / 4
            & \underline{51} / 0
            & 44 / 0
            & 63.2 / 1.6 \\
        starVLA
            & \underline{96} / 38
            & 84 / 0
            & \underline{43} / 8
            & 42 / 0
            & \underline{71} / 2
            & \underline{67.2} / 9.6 \\
        Fast-WAM
            & \textbf{100} / 1
            & \textbf{92} / 0
            & \textbf{64} / 0
            & \textbf{88} / 0
            & \textbf{81} / 0
            & \textbf{85.0} / 0.2 \\
        $\pi_0$
            & \underline{96} / \textbf{80}
            & 37 / 11
            & 27 / 6
            & 28 / 6
            & 17 / 4
            & 41.0 / \underline{21.4} \\
        \midrule
        Ego4Gen
            & 91 / 20
            & 61 / \underline{20}
            & 37 / \underline{13}
            & 26 / \underline{13}
            & 18 / \textbf{10}
            & 46.6 / 15.2 \\
        Ego4Gen (Mixed)
            & 78 / \underline{58}
            & 46 / \textbf{23}
            & 22 / \textbf{21}
            & 27 / \textbf{21}
            & 13 / \underline{9}
            & 37.2 / \textbf{26.4} \\
        \bottomrule
    \end{tabular}
\vspace{-0.4cm}
\end{table*}

To validate Ego4Gen's closed-loop control capabilities, we conduct experiments on RoboTwin 2.0 using the ALOHA-AgileX embodiment. We first post-train Ego4Gen for 10k steps on clean demonstrations of all 50 tasks. The post-trained generator then produces geometry-motion latents for clean demonstrations, which are paired with the corresponding action trajectories for the second-stage training of FlowInverse. During execution, Ego4Gen generates future latents from the current head-camera observation every 40 control steps. FlowInverse then uses these latents to predict actions. We evaluate Ego4Gen on five tasks under the Easy (clean) and Hard (randomized) settings.

We choose Diffusion Policy (DP) \cite{chi2024diffusionpolicy}, TurboVLA \cite{xie2026turbovla}, starVLA \cite{community2026starvla}, Fast-WAM \cite{yuan2026fastwam}, and $\pi_0$ \cite{BlackK-RSS-25} as baselines. All methods are trained only on the clean RoboTwin dataset. As shown in \cref{closed_loop}, the clean-only Ego4Gen policy achieves mean success rates of 46.6\% on Easy and 15.2\% on Hard. Compared with the baselines, it achieves the highest Hard-setting success on Place Empty Cup, Pick Diverse Bottles, Place Shoe, and Place Bread Basket. 

\begin{figure}[t]\centering
	\includegraphics[width=1.0\linewidth, clip=true]{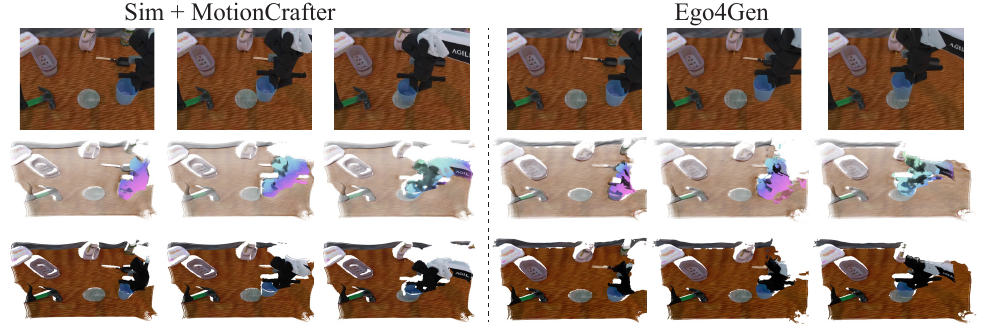}  % left bottom right top
	\caption{Qualitative evaluation of Ego4Gen for closed-loop control. Left: MotionCrafter reconstructions from the simulation video. Right: outputs jointly generated by Ego4Gen. Both sequences start at the same time, and the three image pairs correspond to matching timestamps.}\label{imagination}
\vspace{-0.4cm}
\end{figure}

To further examine whether adapting the world model without additional action supervision can improve policy generalization, we start from the same warmup checkpoint used in the clean-only setting and post-train Ego4Gen on both clean and randomized videos without action labels. We retain the same FlowInverse checkpoints as in the clean-only configuration. With only 5k post-training steps, half as many as the clean-only configuration, Ego4Gen (Mixed) improves Hard-setting success on four of the five tasks, raising the mean from 15.2\% to 26.4\% (\cref{closed_loop}). Mean Easy-setting success decreases from 46.6\% to 37.2\%. We think this decline may partly stem from a mismatch between the updated generator's latents and those used to train FlowInverse. We also visualize Ego4Gen's outputs during closed-loop control in \cref{imagination}. The results show that FlowInverse translates Ego4Gen's joint predictions into robot actions that closely follow the predicted motion over time.

Together, these experiments demonstrate the utility of geometry-motion latents for translating predicted 4D futures into robot control. The gains on randomized scenes with a fixed FlowInverse further show that this interface supports improved policy generalization through world-model adaptation using videos without action labels, suggesting a path toward reduced reliance on action-labeled data.

\subsection{Ablation Study}

\subsubsection{World Model Components}

\begin{table}[t]
    \centering
    \caption{Component ablations on EgoDex test clips (4\,s). AA: asymmetric attention; GI: gradient isolation.}
    \label{wm_ablation}
    \setlength{\tabcolsep}{4pt}
    \begin{tabular}{@{}lcccc@{}}
        \toprule
        Variant & PSNR$\uparrow$ & SSIM$\uparrow$
                & LPIPS$\downarrow$ & NC$\downarrow$ \\
        \midrule
        Shared
            & 20.95 & 0.766 & 0.297 & 0.02382 \\
        Routed only
            & 22.44 & 0.798 & 0.230 & 0.01906 \\
        Routed + AA
            & \underline{23.02}
            & \underline{0.807}
            & \underline{0.209}
            & \textbf{0.01446} \\
        Routed + GI
            & 22.95
            & 0.801
            & 0.212
            & 0.01575 \\
        \midrule
        Full
            & \textbf{23.55}
            & \textbf{0.820}
            & \textbf{0.180}
            & \underline{0.01472} \\
        \bottomrule
    \end{tabular}
\vspace{-0.4cm}
\end{table}

We examine the effects of modality-routed LoRA, asymmetric attention, and gradient isolation by comparing the full model with four variants: (i) shared LoRA, which uses a single set of adapters across all modalities (r=64, $\alpha$=128) while retaining asymmetric attention and gradient isolation; (ii) routed LoRA without either component; (iii) routed LoRA with asymmetric attention only; and (iv) routed LoRA with gradient isolation only.

We train all variants and the full model for 5k steps using the same batch size and learning rate, and evaluate them on the same 50 EgoDex test clips. We assess vision quality using PSNR, SSIM, and LPIPS, and geometric surface smoothness using intra-mesh normal consistency (NC), where lower values indicate smoother surfaces. We omit motion from this comparison because differences across variants are small at this training stage. Results are reported in \cref{wm_ablation}.

These results suggest that combining modality-routed LoRA with asymmetric attention and gradient isolation helps preserve the model's pretrained priors when training data are limited, while enabling more efficient learning of newly introduced modalities.

\subsubsection{FlowInverse Pretraining}

FlowInverse is pretrained on multiple manipulation tasks in both \cref{versatile_representation} and \cref{CLRC}. We hypothesize that this stage provides a useful prior for mapping geometry-motion representations to robot actions. To validate this hypothesis, we remove the 50-task pretraining stage from the clean-only configuration in \cref{CLRC} and train each task-specific FlowInverse model using the same Ego4Gen-generated geometry-motion latents dataset. We then evaluate the policies on the same five tasks under the Easy and Hard settings, with results reported in \cref{flowinverse_pretraining}.

\begin{table}[t]
    \centering
    \caption{Effect of FlowInverse pretraining on closed-loop success rates (\%, $\uparrow$). Each cell reports Easy / Hard.}
    \label{flowinverse_pretraining}
    \setlength{\tabcolsep}{4pt}
    \begin{tabular}{@{}lcc@{}}
        \toprule
        Task & w/o Pretraining & Ego4Gen \\
        \midrule
        Grab Roller
            & \underline{76} / \textbf{31}
            & \textbf{91} / \underline{20} \\
        Place Empty Cup
            & \underline{3} / \underline{0}
            & \textbf{61} / \textbf{20} \\
        Pick Diverse Bottles
            & \underline{8} / \underline{4}
            & \textbf{37} / \textbf{13} \\
        Place Shoe
            & \underline{0} / \underline{0}
            & \textbf{26} / \textbf{13} \\
        Place Bread Basket
            & \underline{5} / \underline{2}
            & \textbf{18} / \textbf{10} \\
        \midrule
        Mean
            & \underline{18.4} / \underline{7.4}
            & \textbf{46.6} / \textbf{15.2} \\
        \bottomrule
    \end{tabular}
\vspace{-0.4cm}
\end{table}

These results show that pretraining on diverse manipulation tasks enables FlowInverse to learn a transferable prior for mapping geometry-motion latents to robot actions, improving its performance under the guidance of the world model.
\section{Conclusion}

We present Ego4Gen, which generates geometry-motion latents as an interface between world prediction and action execution. Given a single egocentric image and a text prompt, Ego4Gen jointly generates temporally aligned future RGB video, dense point maps, and dense scene flow. Together with FlowInverse, a lightweight inverse dynamics network, it enables closed-loop robot control. Experiments validate the utility of geometry-motion representations for both explicit 4D scene reconstruction and robot action prediction, as well as their robustness in out-of-distribution settings. We further show that adapting the world model with videos without action labels improves the mean success rate in randomized scenes while keeping FlowInverse weights fixed. These results support decoupled training of future generation and action mapping through a geometry-motion interface, providing a path to improving downstream control with more action-unlabeled videos and reducing reliance on action-labeled data.

{
    \small
    \bibliographystyle{ieeenat_fullname}
    \bibliography{main}

@String(CVPR= {IEEE Conf. Comput. Vis. Pattern Recog.})

@String(ICCV= {Int. Conf. Comput. Vis.})

@String(ECCV= {Eur. Conf. Comput. Vis.})

@String(CVPR  = {CVPR})

@String(ICCV  = {ICCV})

@String(ECCV  = {ECCV})

@InProceedings{Zhu_2026_CVPR,
    author    = {Zhu, Ruijie and Lu, Jiahao and Hu, Wenbo and Han, Xiaoguang and Cai, Jianfei and Shan, Ying and Zheng, Chuanxia},
    title     = {MotionCrafter: Dense Geometry and Motion Reconstruction with a 4D VAE},
    booktitle = {Proceedings of the IEEE/CVF Conference on Computer Vision and Pattern Recognition (CVPR)},
    month     = {June},
    year      = {2026},
    pages     = {40303-40315}
}

@techreport{nvidia2026cosmos3,
  title  = {Cosmos 3: Omnimodal World Models for physical AI},
  author = {{NVIDIA Cosmos Team}},
  year   = {2026},
  url    = {https://research.nvidia.com/labs/cosmos-lab/cosmos3/technical-report.pdf}
}

@article{wan2025,
      title={Wan: Open and Advanced Large-Scale Video Generative Models}, 
      author={Team Wan and Ang Wang and Baole Ai and Bin Wen and Chaojie Mao and Chen-Wei Xie and Di Chen and Feiwu Yu and Haiming Zhao and Jianxiao Yang and Jianyuan Zeng and Jiayu Wang and Jingfeng Zhang and Jingren Zhou and Jinkai Wang and Jixuan Chen and Kai Zhu and Kang Zhao and Keyu Yan and Lianghua Huang and Mengyang Feng and Ningyi Zhang and Pandeng Li and Pingyu Wu and Ruihang Chu and Ruili Feng and Shiwei Zhang and Siyang Sun and Tao Fang and Tianxing Wang and Tianyi Gui and Tingyu Weng and Tong Shen and Wei Lin and Wei Wang and Wei Wang and Wenmeng Zhou and Wente Wang and Wenting Shen and Wenyuan Yu and Xianzhong Shi and Xiaoming Huang and Xin Xu and Yan Kou and Yangyu Lv and Yifei Li and Yijing Liu and Yiming Wang and Yingya Zhang and Yitong Huang and Yong Li and You Wu and Yu Liu and Yulin Pan and Yun Zheng and Yuntao Hong and Yupeng Shi and Yutong Feng and Zeyinzi Jiang and Zhen Han and Zhi-Fan Wu and Ziyu Liu},
      journal = {arXiv preprint arXiv:2503.20314},
      year={2025}
}

@InProceedings{pmlr-v306-gao26q,
  title = 	 {{D}ream{D}ojo: A Generalist Robot World Model from Large-Scale Human Videos},
  author =       {Gao, Shenyuan and Liang, William and Zheng, Kaiyuan and Malik, Ayaan Naveed and Ye, Seonghyeon and Yu, Sihyun and Tseng, Wei-Cheng and Dong, Yuzhu and Mo, Kaichun and Lin, Chen-Hsuan and Xiang, Jiannan and Xie, Yuqi and Zheng, Ruijie and Niu, Dantong and Jannaty, Pooya and Gu, Jinwei and Zhang, Jun and Malik, Jitendra and Abbeel, Pieter and Liu, Ming-Yu and Zhu, Yuke and Jang, Joel and Fan, Linxi},
  booktitle = 	 {Proceedings of the 43rd International Conference on Machine Learning},
  pages = 	 {33340--33366},
  year = 	 {2026},
  editor = 	 {Zhang, Tong and Dudik, Miroslav and Jaggi, Martin and Agarwal, Alekh and Li, Sharon and Schuurmans, Dale and Zhu, Jerry and Berkenkamp, Felix and Dong, Hanze and Bietti, Alberto},
  volume = 	 {306},
  series = 	 {Proceedings of Machine Learning Research},
  month = 	 {06--11 Jul},
  publisher =    {PMLR},
  url = 	 {https://proceedings.mlr.press/v306/gao26q.html}
}

@INPROCEEDINGS{BuQ-RSS-25, 
    AUTHOR    = {Qingwen Bu AND Yanting Yang AND Jisong Cai AND Shenyuan Gao AND Guanghui Ren AND Maoqing Yao AND Ping Luo AND Hongyang Li}, 
    TITLE     = {{Learning to Act Anywhere with Task-centric Latent Actions}}, 
    BOOKTITLE = {Proceedings of Robotics: Science and Systems}, 
    YEAR      = {2025}, 
    ADDRESS   = {LosAngeles, CA, USA}, 
    MONTH     = {June}, 
    DOI       = {10.15607/RSS.2025.XXI.014} 
}

@article{chen2025robotwin,
  title={Robotwin 2.0: A scalable data generator and benchmark with strong domain randomization for robust bimanual robotic manipulation},
  author={Chen, Tianxing and Chen, Zanxin and Chen, Baijun and Cai, Zijian and Liu, Yibin and Li, Zixuan and Liang, Qiwei and Lin, Xianliang and Ge, Yiheng and Gu, Zhenyu and others},
  journal={arXiv preprint arXiv:2506.18088},
  year={2025}
}

@inproceedings{hoque2026egodex,
  title={Egodex: Learning dexterous manipulation from large-scale egocentric video},
  author={Hoque, Ryan and Huang, Peide and Yoon, David and Zhang, Jian and others},
  booktitle={International Conference on Learning Representations},
  volume={2026},
  pages={4218--4237},
  year={2026}
}

@misc{agibotworld2026,
  title        = {AgiBot World 2026},
  author       = {AgiBot World Team},
  howpublished = {\url{https://huggingface.co/datasets/agibot-world/AgiBotWorld2026}},
  year         = {2026}
}

@InProceedings{Liu_2024_CVPR,
    author    = {Liu, Yun and Yang, Haolin and Si, Xu and Liu, Ling and Li, Zipeng and Zhang, Yuxiang and Liu, Yebin and Yi, Li},
    title     = {TACO: Benchmarking Generalizable Bimanual Tool-ACtion-Object Understanding},
    booktitle = {Proceedings of the IEEE/CVF Conference on Computer Vision and Pattern Recognition (CVPR)},
    month     = {June},
    year      = {2024},
    pages     = {21740-21751}
}

@article{rynnworld4d,
  title  = {RynnWorld-4D: 4D Embodied World Models for Robotic Manipulation},
  author = {Haoyu Zhao and Xingyue Zhao and Siteng Huang and Xin Li and Deli Zhao and Zhongyu Li},
  journal= {arXiv preprint arXiv:2607.06559},
  year   = {2026},
}

@article{chen20254dnex,
    title={4DNeX: Feed-Forward 4D Generative Modeling Made Easy},
    author={Chen, Zhaoxi and Liu, Tianqi and Zhuo, Long and Ren, Jiawei and Tao, Zeng and Zhu, He and Hong, Fangzhou and Pan, Liang and Liu, Ziwei},
    journal={arXiv preprint arXiv:2508.13154},
    year={2025}
}

@InProceedings{Zhen_2025_ICCV,
    author    = {Zhen, Haoyu and Sun, Qiao and Zhang, Hongxin and Li, Junyan and Zhou, Siyuan and Du, Yilun and Gan, Chuang},
    title     = {Learning 4D Embodied World Models},
    booktitle = {Proceedings of the IEEE/CVF International Conference on Computer Vision (ICCV)},
    month     = {October},
    year      = {2025},
    pages     = {5337-5347}
}

@inproceedings{wang26vggt-omega,
 author = {Jianyuan Wang and Minghao Chen and Shangzhan Zhang and Nikita Karaev and Johannes Sch{\"o}nberger and Patrick Labatut and Piotr Bojanowski and David Novotny and Andrea Vedaldi and Christian Rupprecht},
 booktitle = {Proceedings of the {IEEE} Conference on Computer Vision and Pattern Recognition ({CVPR})},
 title = {{VGGT-$\Omega$}},
 year = {2026}
}

@inproceedings{harley2025alltracker,
author    = {Adam W. Harley and Yang You and Xinglong Sun and Yang Zheng and Nikhil Raghuraman and Yunqi Gu and Sheldon Liang and Wen-Hsuan Chu and Achal Dave and Pavel Tokmakov and Suya You and Rares Ambrus and Katerina Fragkiadaki and Leonidas J. Guibas},
title     = {All{T}racker: {E}fficient Dense Point Tracking at High Resolution},
booktitle = {ICCV},
year      = {2025}
}

@article{chi2024diffusionpolicy,
	author = {Cheng Chi and Zhenjia Xu and Siyuan Feng and Eric Cousineau and Yilun Du and Benjamin Burchfiel and Russ Tedrake and Shuran Song},
	title ={Diffusion Policy: Visuomotor Policy Learning via Action Diffusion},
	journal = {The International Journal of Robotics Research},
	year = {2024},
}

@article{xie2026turbovla,
  title  = {{TurboVLA}: Real-Time Vision-Language-Action Model at
            32 {Hz} on an {RTX} 4090 with {$<1$} {GB} {VRAM}},
  author = {Xie, Hengyi and Yao, Chenfei and Wu, Xianjin and
            Xi, Xuanyang and Tang, Yiping and Xu, Di and
            Zhu, Yingying and Liang, Dingkang and Bai, Xiang and
            Ding, Han},
  journal = {arXiv preprint arXiv:2607.27205},
  year   = {2026}
}

@article{community2026starvla,
  title={StarVLA: A Lego-like Codebase for Vision-Language-Action Model Developing},
  author={{StarVLA Community}},
  journal={arXiv preprint arXiv:2604.05014},
  year={2026},
  eprint={2604.05014},
  archivePrefix={arXiv},
  primaryClass={cs.RO}
}

@article{yuan2026fastwam,
  title={Fast-WAM: Do World Action Models Need Test-time Future Imagination?},
  author={Tianyuan Yuan and Zibin Dong and Yicheng Liu and Hang Zhao},
  journal={arXiv preprint arXiv:2603.16666},
  year={2026},
  url={https://arxiv.org/abs/2603.16666}
}

@INPROCEEDINGS{BlackK-RSS-25, 
    AUTHOR    = {Kevin Black AND Noah Brown AND Danny Driess AND Adnan Esmail AND Michael Robert Equi AND Chelsea Finn AND Niccolo Fusai AND Lachy Groom AND Karol Hausman AND Brian Ichter AND Szymon Jakubczak AND Tim Jones AND Liyiming Ke AND Sergey Levine AND Adrian Li-Bell AND Mohith Mothukuri AND Suraj Nair AND Karl Pertsch AND Lucy Xiaoyang Shi AND Laura Smith AND James Tanner AND Quan Vuong AND Anna Walling AND Haohuan Wang AND Ury Zhilinsky}, 
    TITLE     = {{$\pi_0$: A Vision-Language-Action Flow Model for General Robot Control}}, 
    BOOKTITLE = {Proceedings of Robotics: Science and Systems}, 
    YEAR      = {2025}, 
    ADDRESS   = {LosAngeles, CA, USA}, 
    MONTH     = {June}, 
    DOI       = {10.15607/RSS.2025.XXI.010} 
}

@inproceedings{NEURIPS2023_1d5b9233,
 author = {Du, Yilun and Yang, Sherry and Dai, Bo and Dai, Hanjun and Nachum, Ofir and Tenenbaum, Josh and Schuurmans, Dale and Abbeel, Pieter},
 booktitle = {Advances in Neural Information Processing Systems},
 doi = {10.52202/075280-0403},
 editor = {A. Oh and T. Naumann and A. Globerson and K. Saenko and M. Hardt and S. Levine},
 pages = {9156--9172},
 publisher = {Curran Associates, Inc.},
 title = {Learning Universal Policies via Text-Guided Video Generation},
 url = {https://proceedings.neurips.cc/paper_files/paper/2023/file/1d5b9233ad716a43be5c0d3023cb82d0-Paper-Conference.pdf},
 volume = {36},
 year = {2023}
}

@inproceedings{ye2026rss,
  title = {World Action Models are Zero-shot Policies},
  author = {Seonghyeon Ye and Yunhao Ge and Kaiyuan Zheng and Shenyuan Gao and Sihyun Yu and George Kurian and Suneel Indupuru and You Liang Tan and Chuning Zhu and Jiannan Xiang and Ayaan Naveed Malik and Kyungmin Lee and William Liang and Nadun Ranawaka Arachchige and Jiasheng Gu and Yinzhen Xu and Guanzhi Wang and Fengyuan Hu and Avnish Narayan and Johan Bjorck and Jing Wang and Gwanghyun Kim and Dantong Niu and Ruijie Zheng and Yuqi Xie and Jimmy Wu and Qi Wang and Ryan Julian and Danfei Xu and Yilun Du and Yevgen Chebotar and Scott Reed and Jan Kautz and Yuke Zhu and Linxi Fan and Joel Jang},
  year = {2026},
  month = {November},
  booktitle = {Conference on Robot Learning (CoRL)},
  eprint = {2602.15922},
  archiveprefix = {arxiv},
}

@article{assran2025vjepa2,
  title={V-JEPA~2: Self-Supervised Video Models Enable Understanding, Prediction and Planning},
  author={Assran, Mahmoud and Bardes, Adrien and Fan, David and Garrido, Quentin and Howes, Russell and
Komeili, Mojtaba and Muckley, Matthew and Rizvi, Ammar and Roberts, Claire and Sinha, Koustuv and Zholus, Artem and
Arnaud, Sergio and Gejji, Abha and Martin, Ada and Robert Hogan, Francois and Dugas, Daniel and
Bojanowski, Piotr and Khalidov, Vasil and Labatut, Patrick and Massa, Francisco and Szafraniec, Marc and
Krishnakumar, Kapil and Li, Yong and Ma, Xiaodong and Chandar, Sarath and Meier, Franziska and LeCun, Yann and
Rabbat, Michael and Ballas, Nicolas},
  journal={arXiv preprint arXiv:2506.09985},
  year={2025}
}

@inproceedings{NEURIPS2025_d7b3cef7,
 author = {Zhang, Chuheng and Pearce, Tim and Zhang, Pushi and Wang, Kaixin and Chen, Xiaoyu and Shen, Wei and Zhao, Li and Bian, Jiang},
 booktitle = {Advances in Neural Information Processing Systems},
 doi = {10.52202/085713-4906},
 editor = {D. Belgrave and C. Zhang and H. Lin and R. Pascanu and P. Koniusz and M. Ghassemi and N. Chen},
 pages = {146676--146697},
 publisher = {Curran Associates, Inc.},
 title = {What Do Latent Action Models Actually Learn?},
 url = {https://proceedings.neurips.cc/paper_files/paper/2025/file/d7b3cef7c31b94a4a533db83d01a8882-Paper-Conference.pdf},
 volume = {38, Main Conference},
 year = {2025}
}

@InProceedings{pmlr-v270-xu25a,
  title = 	 {Flow as the Cross-domain Manipulation Interface},
  author =       {Xu, Mengda and Xu, Zhenjia and Xu, Yinghao and Chi, Cheng and Wetzstein, Gordon and Veloso, Manuela and Song, Shuran},
  booktitle = 	 {Proceedings of The 8th Conference on Robot Learning},
  pages = 	 {2475--2499},
  year = 	 {2025},
  editor = 	 {Agrawal, Pulkit and Kroemer, Oliver and Burgard, Wolfram},
  volume = 	 {270},
  series = 	 {Proceedings of Machine Learning Research},
  month = 	 {06--09 Nov},
  publisher =    {PMLR},
  url = 	 {https://proceedings.mlr.press/v270/xu25a.html}
}

@misc{zhi20253dflowactionlearningcrossembodimentmanipulation,
      title={3DFlowAction: Learning Cross-Embodiment Manipulation from 3D Flow World Model}, 
      author={Hongyan Zhi and Peihao Chen and Siyuan Zhou and Yubo Dong and Quanxi Wu and Lei Han and Mingkui Tan},
      year={2025},
      eprint={2506.06199},
      archivePrefix={arXiv},
      primaryClass={cs.RO},
      url={https://arxiv.org/abs/2506.06199}, 
}

@misc{li_hydra-0_2026,
    title = {Hydra-0: {Action} {Flow} for {Generalist} {World} {Modeling} and {Control}},
    shorttitle = {Hydra-0},
    url = {http://arxiv.org/abs/2608.18077},
    doi = {10.48550/arXiv.2608.18077},
    publisher = {arXiv},
    author = {Li, Hongyu and Wen, Bowen and Zhu, Xinghao and Wang, Yixuan and Du, Yilun and Li, Yunzhu and Konidaris, George and Birchfield, Stan and Pouya, Soha and Li, Chenran and Chang, Yan},
    month = aug,
    year = {2026},
    note = {arXiv:2608.18077 [cs.RO]},
}

@misc{sun2026riemann10embodiedworldaction,
      title={Riemann-1.0: An Embodied World Action Model for Physical AI}, 
      author={Haofeng Sun and Jiangbo Pei and Fei Kang and Zexiang Liu and Yaokun Li and Boyi Jiang and Hua Xue and Cindy Zhou and Wei Li and Yichen Wei and Mengyin An and Fanliang Zhao and Biao Jiang and Zile Wang and Yang Liu and Yangguang Li},
      year={2026},
      eprint={2608.27033},
      archivePrefix={arXiv},
      primaryClass={cs.RO},
      url={https://arxiv.org/abs/2608.27033}, 
}

@misc{guo2026unified4dworldaction,
      title={Unified 4D World Action Modeling from Video Priors with Asynchronous Denoising}, 
      author={Jun Guo and Qiwei Li and Peiyan Li and Zilong Chen and Nan Sun and Yifei Su and Heyun Wang and Yuan Zhang and Xinghang Li and Huaping Liu},
      year={2026},
      eprint={2604.26694},
      archivePrefix={arXiv},
      primaryClass={cs.RO},
      url={https://arxiv.org/abs/2604.26694}, 
}

@inproceedings{mi2026one4d,
  title={One4D: Unified 4D Generation and Reconstruction via Decoupled LoRA Control},
  author={Mi, Zhenxing and Wang, Yuxin and Xu, Dan},
  booktitle={European Conference on Computer Vision (ECCV)},
  year={2026}
}

@misc{liu2026pixelsvideopriors4d,
  title         = {Beyond Pixels: From Video Priors to 4D Worlds},
  author        = {Zihao Liu and Xiaolong Shen and Zhenglin Zhou and Ruijie Quan and Yi Yang},
  year          = {2026},
  eprint        = {2608.10744},
  archivePrefix = {arXiv},
  primaryClass  = {cs.CV},
  url           = {https://arxiv.org/abs/2608.10744}
}

@article{lu2026track4world,
    title   = {Track4World: Feedforward World-Centric Dense 3D Tracking of All Pixels},
    author  = {Jiahao Lu and Jiayi Xu and Wenbo Hu and Ruijie Zhu and Chengfeng Zhao and Sai-Kit Yeung and Ying Shan and Yuan Liu},
    journal = {arXiv preprint arXiv:2603.02573},
    year    = {2026}
}

@article{depthanything3,
  title={Depth Anything 3: Recovering the visual space from any views},
  author={Haotong Lin and Sili Chen and Jun Hao Liew and Donny Y. Chen and Zhenyu Li and Guang Shi and Jiashi Feng and Bingyi Kang},
  journal={arXiv preprint arXiv:2511.10647},
  year={2025}
}
}

\clearpage
\setcounter{page}{1}
% \maketitlesupplementary

\twocolumn[{%
  \begin{minipage}{\textwidth}
    \centering
    {\Large\bfseries \thetitle\par}
    \vspace{0.5em}
    {\Large Supplementary Material\par}
    \vspace{0.8em}

    \includegraphics[width=\linewidth]{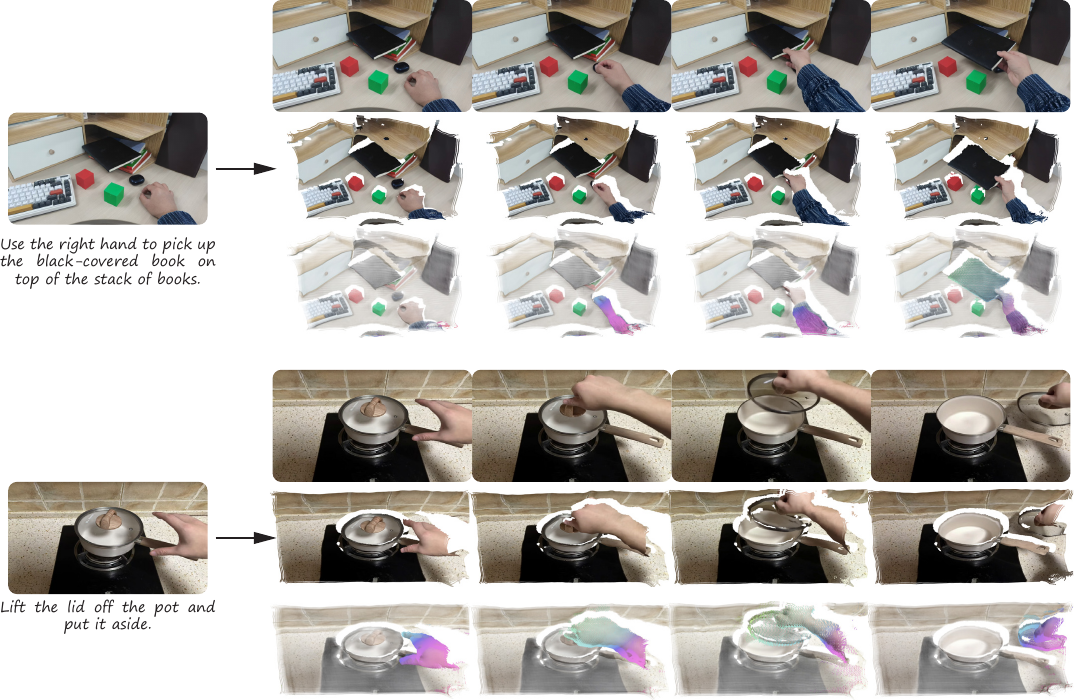}\par
    \captionof{figure}{Additional examples of 4D generation by Ego4Gen. The leftmost column shows the input image and instruction. Subsequent columns show the predictions in chronological order from left to right.}
    \label{show}
  \end{minipage}\par
  \vspace{\dbltextfloatsep}
}]

\section{Implementation Details}

\paragraph{Model Training.}
For Ego4Gen, we freeze the Cosmos3-Nano backbone and add LoRA adapters to the query, key, value, and output projections in the generator's attention layers. We set the LoRA rank and scaling parameter to $(r,\alpha)=(32,64)$ for the vision branch and $(64,128)$ for both the geometry and motion branches. During RoboTwin post-training, the Clean variant uses 50 clean demonstrations for each of the 50 tasks. The Mixed variant adds 500 videos from randomized scenes per task.

To construct the adaptation data for FlowInverse, we sample a head-camera image every four control steps from each expert demonstration. Given the image and task instruction, the frozen world model Ego4Gen generates four seconds of geometry-motion latents in 10 denoising steps. We pair these latents with expert actions at the corresponding timestamps in the same trajectory. FlowInverse encodes the geometry and motion latents with separate convolutional branches. It is a three-layer attention decoder with six heads and a hidden dimension of 192. It combines these features with the current robot states to predict action waypoints over the next two seconds. At deployment, the robot runs at 15\,Hz: FlowInverse updates the action plan every eight control steps, and Ego4Gen regenerates the geometry-motion latents every 40 steps.

\paragraph{4D Generation Evaluation.}
We evaluate all models on a fixed set of TACO clips using the same initial RGB frame and task instructions for each clip. We score predictions at shared future timestamps. Ego4Gen uses 35 denoising steps for this evaluation. For point map evaluation, we fit a single similarity transform on the first frame and keep it fixed for all subsequent frames. To obtain depth from Ego4Gen's aligned point maps, we use TACO's ground-truth camera extrinsics to transform the points into each frame's camera coordinates and take their $z$-coordinates. We then align each depth map to the reference in scale and shift. Because 4DNeX centers its point maps independently in each frame, we additionally use a per-frame perspective-$n$-point (PnP) alignment. And we also show some 4D generation examples of Ego4Gen in \cref{show}.

\section{Additional Experiments}

\subsection{Pretraining on Non-Target Tasks}

In \cref{versatile_representation}, we pretrain FlowInverse for each representation on data from 48 non-target RoboTwin tasks. The purpose of this pretraining is to learn a latent-to-action prior from a large set of demonstrations. To evaluate its effect, we omit the 48-task pretraining stage for each representation and train a separate decoder directly on each target task. \Cref{non_target_pretraining} reports the resulting success rates.

\begin{table}[t]
    \centering
    \caption{Effect of pretraining on 48 non-target tasks. Each cell reports success rates (\%) on evaluation scenes, without $\to$ with pretraining.}
    \label{non_target_pretraining}
    \small
    \setlength{\tabcolsep}{2.5pt}
    \begin{tabular}{@{}lcccc@{}}
        \toprule
        \multirow{2}{*}{Representation}
            & \multicolumn{2}{c}{Place Empty Cup}
            & \multicolumn{2}{c}{Pick Diverse Bottles} \\
        \cmidrule(lr){2-3}\cmidrule(lr){4-5}
            & Easy & Hard & Easy & Hard \\
        \midrule
        GM
            & $30\!\to\!82$ & $13\!\to\!51$
            & $26\!\to\!46$ & $19\!\to\!37$ \\
        M-only
            & $33\!\to\!60$ & $18\!\to\!55$
            & $15\!\to\!31$ & $13\!\to\!26$ \\
        \midrule
        DreamDojo
            & $48\!\to\!88$ & $1\!\to\!0$
            & $35\!\to\!35$ & $0\!\to\!0$ \\
        UniVLA
            & $20\!\to\!51$ & $0\!\to\!0$
            & $5\!\to\!12$  & $1\!\to\!4$ \\
        Wan2.2
            & $2\!\to\!16$  & $0\!\to\!0$
            & $9\!\to\!31$  & $0\!\to\!0$ \\
        \bottomrule
    \end{tabular}
\end{table}

As shown in \cref{non_target_pretraining}, pretraining improves success in 14 of the 20 comparisons. Both GM and M-only benefit on both tasks under Easy and Hard conditions, whereas gains for the other representations occur mainly under Easy conditions. These results suggest that the mapping from geometry-motion latents to actions transfers across manipulation tasks.

\subsection{Closed-Loop Control with Different Representations}

We next test whether the trends in \cref{versatile_representation} extend from expert futures to futures predicted during closed-loop execution. We encode clean RoboTwin demonstrations using MotionCrafter-generated geometry-motion latents, the DreamDojo or UniVLA LAM, or the Wan2.2 VAE. For each representation, we pretrain a FlowInverse decoder on all 50 tasks with a budget comparable to that of our main policy in \cref{closed_loop}. We then fine-tune each decoder on its target task by pairing representations extracted from Ego4Gen-generated RGB videos with the corresponding expert actions. At deployment, the Clean variant of Ego4Gen predicts future RGB frames from the current observation and task instruction using 10 denoising steps. Each baseline encodes these frames for control, while our main policy uses the jointly generated geometry-motion latents directly.

\begin{table}[t]
    \centering
    \caption{Closed-loop success rates (\%, $\uparrow$) with native geometry-motion latents versus representations extracted from predicted RGB frames. The Ego4Gen row reports the Clean policy from \cref{closed_loop}.}
    \label{closed_loop_representations}
    \small
    \setlength{\tabcolsep}{2.5pt}
    \begin{tabular}{@{}lcccc@{}}
        \toprule
        \multirow{2}{*}{Representation}
            & \multicolumn{2}{c}{Place Empty Cup}
            & \multicolumn{2}{c}{Pick Diverse Bottles} \\
        \cmidrule(lr){2-3}\cmidrule(lr){4-5}
            & Easy & Hard & Easy & Hard \\
        \midrule
        Ego4Gen
            & \textbf{61} & \textbf{20}
            & \textbf{37} & \textbf{13} \\
        \midrule
        MotionCrafter
            & 49 & \underline{15} & \underline{32} & \underline{11} \\
        DreamDojo & 19 & 1 & 18 & 1 \\
        UniVLA    &  6 & 1 &  2 & 1 \\
        Wan2.2 VAE    & \underline{51} & 0 & 23 & 1 \\
        \bottomrule
    \end{tabular}
\end{table}

The results show that Ego4Gen's native geometry-motion latents yield higher success rates in all four settings than the geometry-motion latents extracted by MotionCrafter from predicted RGB. The comparison with DreamDojo, UniVLA, and Wan2.2 further demonstrates the advantage of the geometry-motion representation for closed-loop control under distribution shift.

\end{document}